# Advanced Clustering Framework for Semiconductor Image Analytics Integrating Deep TDA with Self-Supervised and Transfer Learning Techniques

Janhavi Giri[1*,] Attila Lengyel[1], Don Kent[1], Edward Kibardin[2]

[1] Intel Corporation, Santa Clara, CA, USA

[2] DataRefiner, London, UK

*Correspondence: janhavi.giri@intel.com

## Abstract

Semiconductor manufacturing generates vast amounts of image data, crucial for defect identification and yield optimization, yet often exceeds manual inspection capabilities. Traditional clustering techniques struggle with high-dimensional, unlabeled data, limiting their effectiveness in capturing nuanced patterns. This paper introduces an advanced clustering framework that integrates deep Topological Data Analysis (TDA) with self-supervised and transfer learning techniques, offering a novel approach to unsupervised image clustering. TDA captures intrinsic topological features, while self-supervised learning extracts meaningful representations from unlabeled data, reducing reliance on labeled datasets. Transfer learning enhances the framework's adaptability and scalability, allowing fine-tuning to new datasets without retraining from scratch. Validated on synthetic and open-source semiconductor image datasets, the framework successfully identifies clusters aligned with defect patterns and process variations. This study highlights the transformative potential of combining TDA, self-supervised learning, and transfer learning, providing a scalable solution for proactive process monitoring and quality control in semiconductor manufacturing and other domains with large-scale image datasets.

## Introduction

The semiconductor industry's rapid growth underscores its pivotal role in driving technological progress across diverse fields (1). Semiconductor manufacturing is a cornerstone of modern technology, with global sales projected to reach $697 billion in 2025, driven by advancements in AI, 5G, and generative AI chips (2). This intricate process generates vast and ever-increasing volumes of data, particularly high-resolution images used for monitoring quality control and optimizing production yield (3-5). Identifying minuscule defects and subtle process variations from these images is critical for maintaining high yields and ensuring device reliability. However, the sheer scale and complexity of this image data overwhelm traditional manual inspection methods, which are

often slow, costly, subjective, and infeasible for comprehensive analysis (6-8). Consequently, automated analysis techniques are essential for extracting actionable insights from these image datasets.

Artificial Intelligence (AI), especially Machine Learning (ML) and Deep Learning (DL), has become indispensable in this domain (47, 37), providing powerful tools for extracting significant patterns, detecting subtle anomalies, and advancing automation towards smarter manufacturing. Initial AI adoption heavily relied on supervised learning, particularly Convolutional Neural Networks (CNNs), for defect segmentation and classification where labeled data was available (35, 46). While effective for predefined defects (36), the supervised approach faces significant hurdles: the high cost and effort of labeling data for all possible defects, difficulty managing novel (zero-day) patterns, and challenges in adapting models to process changes, especially with imbalanced data (60).

These limitations spurred research into unsupervised and semi-supervised methods to reduce label dependency. Autoencoders (AEs), Variational Autoencoders (VAEs), and Generative Adversarial Networks (GANs) have been employed for representation learning, anomaly detection, and data augmentation (45, 48, 49, 56, 59, 62). Semi-supervised learning offers a middle ground, combining limited labels with abundant unlabeled data using techniques like self-training (57, 61, 68). Unsupervised clustering aims to discover patterns directly (50, 66), while unsupervised anomaly detection focuses on identifying outliers (55, 58, 64, 65). However, conventional clustering algorithms (e.g., k-means, DBSCAN) often struggle with the high-dimensional, noisy, and complex nature of manufacturing images, failing to capture intricate structures or non-globular patterns effectively (9-16, 43).

Self-Supervised Learning (SSL) has emerged as a particularly promising direction, learning rich representations from unlabeled data via pretext tasks, significantly reducing annotation needs (20-25). Contrastive learning frameworks like momentum contrast have shown strong performance in classifying wafer map defects even with limited data (40-42). SSL's ability to capture underlying data semantics provides a robust foundation for downstream tasks (41, 42, 44).

Complementary to feature learning, Topological Data Analysis (TDA) offers methods for analyzing the shape and structure of data, capturing persistent topological features across multiple scales (17-19,78). While conceptually explored for manufacturing data (19, 71-74), its integration into deep learning frameworks for this specific application remains relatively nascent. Furthermore, Transfer Learning (TL) is crucial for practical deployment, enabling models pre-trained on large datasets to be adapted efficiently to new tasks or domains (26-34, 67), though often implicitly used rather than explicitly integrated within comprehensive unsupervised frameworks. Practical implementation also faces challenges like data drift, scalability, interpretability (54), system integration, and robustness (55, 67).

Despite significant advancements in individual techniques like SSL, a gap exists in developing comprehensive unsupervised frameworks that synergistically integrate the strengths of multiple cutting-edge approaches. The potential benefits of combining SSL's powerful feature learning, Deep TDA's structural sensitivity, and TL's adaptability for robust, label-efficient unsupervised clustering in semiconductor image analytics remain largely underexplored. Current methods may excel in one area but lack the combined strengths needed for optimal pattern discovery and practical deployment.

Therefore, this paper proposes an advanced unsupervised clustering framework that explicitly integrates Deep TDA, SSL, and TL. We hypothesize that this novel synthesis can overcome limitations of existing methods, offering a more robust, adaptable, and insightful solution for unsupervised pattern discovery in complex semiconductor image datasets. We validate this framework on synthetic and real-world open-source datasets, demonstrating its ability to identify distinct clusters corresponding to known defect patterns and process variations, paving the way for enhanced automated visual inspection and quality control.

**Research Objectives**

The primary goal of this research is to implement and validate an advanced, unsupervised clustering framework tailored for the analysis of large-scale semiconductor manufacturing image datasets. This work aims to address critical limitations inherent in traditional clustering methods and supervised approaches when applied to complex, high-dimensional, and often unlabeled industrial image data.

The key objectives are:

- **Unsupervised Clustering using Novel Framework in DataRefiner ©:** To apply the novel framework for unsupervised image clustering in semiconductor manufacturing use cases, utilizing the DataRefiner © approach that integrates Deep Topological Data Analysis (Deep TDA), Self-Supervised Learning (SSL), and Transfer Learning (TL).
- **Reduce Reliance on Labeled Data:** Leveraging SSL to learn rich, discriminative feature representations directly from unlabeled image data, thereby mitigating the significant cost and effort associated with manual annotation in semiconductor manufacturing.
- **Enhance Pattern Discovery with Topology:** Incorporating Deep TDA to capture intrinsic topological and geometric structures within the image data, aiming to improve the separation of clusters based on underlying shapes and connectivity, beyond what pixel-based or standard deep features might capture alone.

- **Improve Adaptability and Efficiency via Transfer Learning:** Utilizing TL by pre-training a foundational model on large, diverse datasets (incorporating both SSL and Deep TDA learning) that can be applied effectively for zero-shot clustering on new datasets or fine-tuned with minimal effort, enhancing deployment speed and scalability. To also develop a distilled, computationally efficient model variant for resource-constrained environments.
- **Address Limitations of Conventional Clustering:** To overcome challenges faced by traditional algorithms (e.g., sensitivity to noise, high dimensionality, difficulty with non-globular shapes) by performing clustering in a robust, learned embedding space shaped by SSL and TDA.
- **Validate Framework Effectiveness:** To demonstrate the framework's ability to identify meaningful clusters corresponding to known defect patterns (single and mixed types) and process variations using both established open-source benchmarks (WM811K, Mixed WM38) and controlled synthetic datasets (SPVD, SWED, as described in Datasets section).

By achieving these objectives, this research seeks to provide a more powerful and practical approach for automated visual inspection in semiconductor manufacturing. The anticipated impact includes:

- **Improved Quality Control:** Enabling proactive identification of subtle defect patterns and process drifts that might be missed by existing methods.
- **Reduced Labor and Cost:** Automating the analysis of vast image datasets, reducing the need for manual inspection and costly data labeling efforts.
- **Faster Root Cause Analysis:** Providing more insightful data clustering that can accelerate the diagnosis of manufacturing issues and prevent yield excursions.
- **Enhanced Efficiency:** Offering scalable solutions through transfer learning and distilled models, suitable for various deployment scenarios from large-scale analysis to edge applications.

Ultimately, this work aims to contribute to a robust, label-efficient, and adaptable method for data-driven decision-making, enhancing yield and quality in semiconductor fabrication and potentially other domains with similar image analysis challenges.

**Methods**

This section outlines the technical methodology underpinning the advanced unsupervised clustering framework. The core approach, conceived and implemented entirely by DataRefiner, integrates deep learning techniques with Topological Data Analysis (TDA) to achieve robust representation learning from image data without reliance on manual labels.

This methodology is enhanced by Transfer Learning (TL) strategies for improved adaptability and deployment efficiency. The implementation leverages standard deep learning libraries such as PyTorch, coupled with DataRefiner's proprietary algorithms and codebases. The multi-stage process is conceptually illustrated in *Figure 1*.

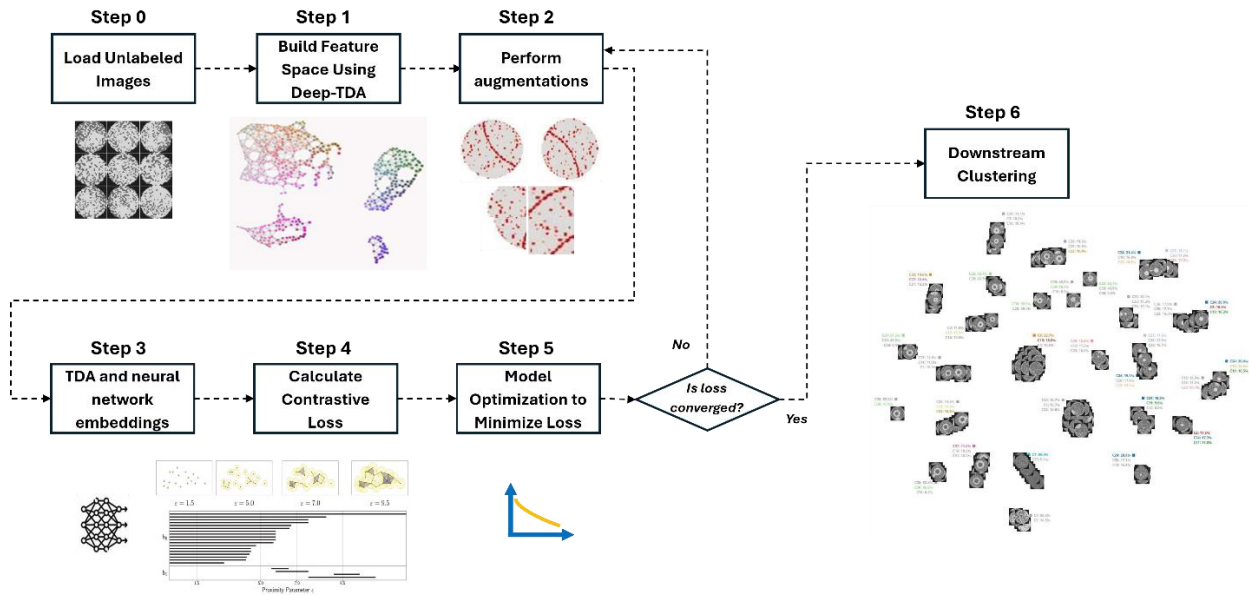

Figure 1 Synergetic framework of Deep Topological Data Analysis (Deep-TDA) and Self-Supervised Learning (SSL) in DataRefiner © enabling characterization of discriminative image representations without the need for labels.

**Core Representation Learning Pipeline: Integrated Deep TDA and Self-Supervised Learning**

The primary learning engine combines Self-Supervised Learning (SSL) with Deep Topological Data Analysis (Deep TDA) to generate discriminative embeddings sensitive to both fine-grained visual patterns and global structural characteristics inherent in the image data.

1. **Deep Topological Feature Extraction:** The process initiates with the extraction of intrinsic topological features from the input image data.

    o *Data Representation:* Input images are transformed into formats amenable to topological analysis (e.g., point clouds based on intensity or feature coordinates).

    o *Topological Computation:* TDA solver computes Persistent Homology (PH) across multiple scales, capturing robust topological invariants (e.g., connected components, loops). These are then vectorized into stable TDA

feature representations (e.g., persistence landscapes or images). These vectors quantify inherent structural properties and serve as direct inputs for the embedding network.

2. **Stochastic Data Augmentation for Contrastive Learning:** To facilitate self-supervision via contrastive learning, multiple augmented views are generated for each input data point (image + its corresponding TDA features).

    - *Augmentation Strategy:* A pipeline of stochastic data augmentations, including geometric and appearance transformations relevant to the image domain, is applied to generate diverse views of the same underlying image content.

    - *Positive Pair Generation:* For each original input, two distinct augmented views are created. These views, along with their associated TDA features, form the positive pairs essential for the contrastive loss function.

3. **Integrated TDA and Neural Network Embedding Generation:** A deep neural network generates low-dimensional vector embeddings that fuse visual information with the pre-computed topological features.

    - *Network Architecture:* A deep Convolutional Neural Network (CNN), such as architectures from the ResNet or EfficientNet families, serves as the visual feature extractor. This network is typically initialized using weights pre-trained on large datasets (Transfer Learning).

    - *Feature Integration:* The topological feature vectors computed in Step 1 TDA methods are explicitly integrated with the visual features from the CNN backbone (e.g., via concatenation) before being processed by a projection head. This ensures the final embedding reflects both visual content and structural topology.

    - *Projection Head:* A small Multi-Layer Perceptron (MLP) maps the combined high-dimensional features into a lower-dimensional embedding space where the contrastive loss is applied.

4. **Contrastive Loss Calculation:** The network is trained using a contrastive loss function (e.g., NT-Xent) to optimize the embedding space.

    - *Objective:* The loss function encourages embeddings from augmented views of the same image (positive pairs) to be similar, while pushing embeddings from different images (negative pairs) apart, typically based on cosine similarity. This optimization guides the network to learn representations

invariant to augmentations but discriminative of semantic and structural content.

5. **Model Optimization:** The parameters of the entire embedding network are iteratively updated via backpropagation using standard gradient-based optimizers (e.g., Adam, SGD). Training proceeds over multiple epochs until the contrastive loss converges, indicating stable and discriminative learned representations.

6. **Downstream Clustering:** After the self-supervised training phase, the projection head is typically removed. The trained encoder (CNN backbone with integrated TDA features) is used to generate final embeddings for the dataset.

    - *Embedding Generation:* Each input image is processed by the trained encoder to produce its low-dimensional embedding vector.

    - *Clustering Application:* DataRefiner utilizes an advanced TDA-enhanced density-based clustering algorithm, developed in-house, applied to these learned embeddings. This algorithm partitions the data into meaningful clusters reflecting inherent patterns, such as different defect classes or process variations. Other standard clustering algorithms could potentially be applied to these embeddings as well.

**Method Extension: Transfer Learning for Generalization and Efficiency**

To enhance the framework's applicability and efficiency, Transfer Learning (TL) is incorporated, leveraging models and techniques discussed in the Methods section.

1. **Pre-training a Foundational Model:** A powerful foundational embedding model is created by applying the integrated SSL and Deep TDA training methodology (Steps 1-5) to a very large and diverse image dataset.

    - *Training Scale:* This pre-training requires significant computational resources (e.g., weeks on high-performance GPUs like NVIDIA A100s) to ensure the model learns generalizable visual and structural features.

    - *Enhanced Generalization:* The explicit integration of Deep TDA component during large-scale pre-training is crucial for capturing fundamental shape and structure characteristics, improving generalization to novel image types beyond what standard CNN features might achieve alone.

2. **Zero-Shot Feature Extraction and Clustering:** The pre-trained foundational model serves as a potent, ready-to-use feature extractor.

- *Inference:* New images can be passed through the pre-trained encoder (CNN + TDA integration) to generate high-quality embeddings without task-specific fine-tuning.
- *Application:* These embeddings can be directly used for downstream tasks like unsupervised clustering (TDA-based algorithm or other methods), enabling rapid analysis and pattern discovery on new datasets (zero-shot learning).

3. **Distilled Model for Resource-Constrained Deployment:** To facilitate deployment in environments lacking high-end GPU resources, a computationally efficient distilled model is created.
    - *Knowledge Distillation:* Techniques are employed to transfer the knowledge from the large "teacher" model to a smaller "student" model with a lightweight architecture (e.g., MobileNetV3).
    - *Efficiency:* The resulting distilled model offers significantly faster inference on CPUs with a smaller memory footprint, providing a practical solution for broader deployment, albeit potentially with a slight trade-off in representational power compared to the full foundational model.

**Datasets**

To rigorously evaluate the performance and versatility of the proposed integrated framework, we utilized a combination of established open-source wafer map datasets and custom-generated synthetic datasets. This section details the characteristics of each dataset employed in our experiments.

*Open source Datasets*

For demonstrating the implementation of the proposed framework, we have leveraged two open-source wafer map datasets: WM811K (75) and Mixed WM38 (76). The open-source datasets comprise of the images generated during the wafer testing phase where the electrical characteristics of each die on the wafer are evaluated, and clusters of defective dies are then presented, creating a distinct spatial pattern known as a wafer map. Wafer defect patterns offer valuable insights for examining problems within the manufacturing process. The defect shapes on the wafer map align with specific wafer defect patterns, allowing for defect identification by observing the wafer map. Consequently, identifying wafer defects and understanding their patterns are crucial for enhancing qualification rates and refining manufacturing procedures.

*WM811K dataset*

The WM811K dataset has become a benchmark for machine learning in semiconductor defect analysis. It is a large collection of semiconductor wafer map images used for failure pattern analysis in manufacturing. It comprises of real failure patterns allowing researchers to train and evaluate algorithms that automatically classify wafers by defect type. WM811K enables the development of automated solutions which has direct industry impact i.e.– fabs can quickly identify yield excursions and trace back root causes without waiting for extensive human analysis, thereby reducing downtime and scrap. In academic research, WM811K is widely used for training and testing defect pattern classification models. It supports a variety of learning scenarios: Supervised classification, Imbalanced learning and augmentation, Semi-supervised and anomaly detection, multi-label pattern recognition, and even Industrial data generation. Its scale and diversity (including noise and variations in wafer layout) provide a robust testbed that mirrors challenges in real fabs (such as noisy data and subtle pattern distinctions). Thus, solutions proven on WM811K can potentially be transferred to actual manufacturing diagnostic systems.

The WM811K dataset contains 811,457 wafer maps collected from about 46,393 production lots in a real-world fabrication environment. Each wafer map is a matrix representing a silicon wafer, where each cell (die) is marked as functional or failed. Out of the 811k wafer maps, 172,950 images have been labeled by domain experts with a failure pattern category. The remaining ~80% of the maps have no specific pattern label (they are either normal or were left unlabeled). The labeled subset is further divided into a training vs. test split as provided by the dataset (with a field indicating "Training" or "Test" for labeled wafers). Each labeled wafer map is assigned to one of nine possible labels based on its pattern of failing dies. These include eight distinct defect pattern classes and one "no pattern" class for normal/no-defect maps. The defect categories are commonly listed as:

1. Center – fails concentrated in the center of the wafer
2. Donut – a ring-shaped failure pattern (with good dies in the center)
3. Edge-Loc – localized fails near the wafer's edge
4. Edge-Ring – a ring of fails along the wafer edge
5. Loc (Local) – a localized cluster of fails (not center or edge)
6. Random – scattered fails with no obvious pattern
7. Scratch – a linear streak of fails (e.g. due to a scratch)
8. Near-full – nearly the entire wafer is failing (almost all dies bad)
9. None (or no pattern) – no specific failure pattern (essentially a normal wafer or random small defects)

Notably, the "none" class dominates the labels – about 85% of the labeled wafers were tagged as having no significant pattern. The remaining 15% (25,519 images) are spread across the eight defect types as shown in Figure 2, making the data highly imbalanced (for example, there are only a few hundred "Donut" examples but tens of thousands of "Center" or "Edge-Ring").

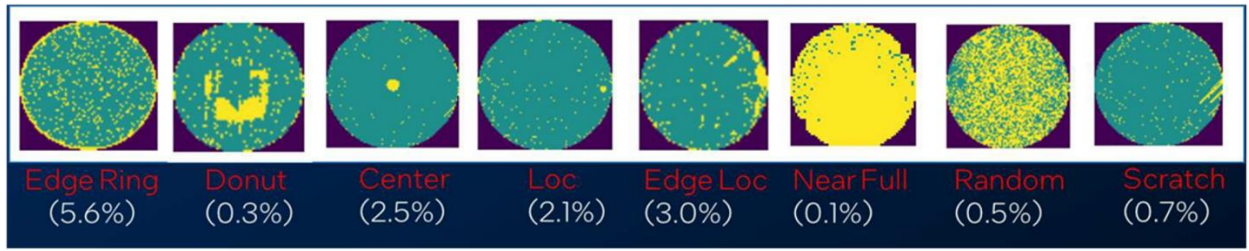

Figure 2 WM811K Open-source wafer map dataset with 811,457 wafer maps from 46,393 lots in real world fabrication. Count of labelled images: 172,950. Count of images with patterns: 25,519. The 8 kinds of patterns and their % distribution shown. 'None' class that contains no defects not shown.

*Mixed WM38 dataset*

Mixed-WM38 is a public dataset of semiconductor wafer map images dedicated to wafer defect pattern recognition, with a special emphasis on mixed-type defects. The Mixed-WM38 dataset was introduced by researchers in 2020 to provide a comprehensive benchmark for detecting multiple concurrent defect patterns on a single wafer. Prior wafer map datasets (like the widely used WM811K) contained only single-pattern defects and normal wafers, lacking examples of mixed-pattern anomalies. The dataset was collected from real production fabs and supplemented with simulated data (using generative models) to ensure a balanced representation of various defect types (77).

Mixed-WM38 contains 38,015 wafer map images in total. Each image (wafer map) is stored as a 52×52 matrix (numpy array) representing the layout of dies on the wafer. The pixel values in a wafer map are encoded as 0, 1, or 2, where **0** denotes areas with no chip (empty/edge of wafer), **1** represents a functioning (good) die that passed testing, and **2** represents a failed (defective) die. Visually, the wafer maps are often displayed with different colors or intensities for 1 vs. 2 to reveal defect patterns (as seen in the figure below). Each wafer's data is accompanied by a label indicating which defect pattern(s) are present.

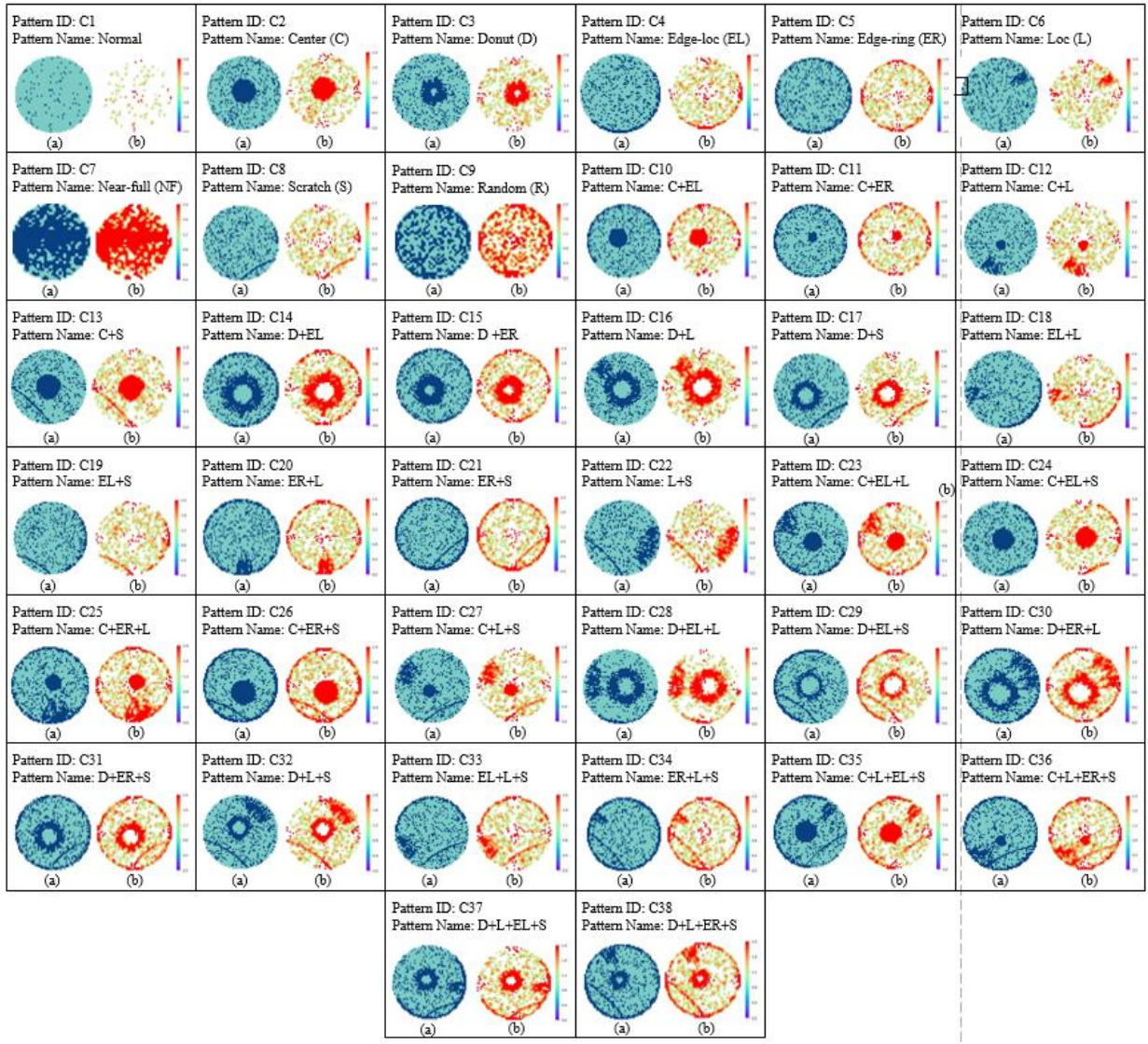

Figure 3 Samples of wafer maps from the Mixed-WM38 dataset, illustrating the range of defect patterns. The top row shows a normal wafer (all dies pass) and eight single-pattern defect types (Center, Donut, Edge-Local, Edge-Ring, Local, Near-Full, Scratch, Random). Subsequent rows show examples of mixed-type defect patterns (e.g. combinations like Center+Edge-Ring, Donut+Scratch, etc.), highlighting multiple failure patterns occurring on one wafer (77).

The dataset encompasses 38 distinct defect categories in total. This set includes:

1. Normal (No Defect): 1 category for defect-free wafers (often labeled "C1" for class 1).
2. Single-Type Defects: 8 categories of a single defect pattern on the wafer (these correspond to the eight fundamental wafer map failure patterns). The single defect classes are typically named: Center (C), Donut (D), Edge-Local (EL), Edge-Ring (ER), Local (Loc), Near-Full (NF), Scratch (S), and Random (R). Each of these reflects a characteristic spatial distribution of failing dies (for example, Center means dies near the wafer center fail, Scratch means a linear streak of fails, Random means scattered fails with no clear pattern, etc.).

3. Mixed-Type Defects: 29 categories of mixed defects, each representing a unique combination of two, three, or four of the above defect patterns occurring together on one wafer.  For example, a wafer exhibiting both a center failure pattern and an edge ring pattern simultaneously would fall into a "Center + Edge-Ring" mixed category. All possible combinations (without repetition) of the 8 base patterns are represented: 13 two-pattern combos, 12 three-pattern combos, and 4 four-pattern combos, totaling 29 mixed classes.  Each combination category is typically denoted by concatenating the abbreviations of its constituent patterns (e.g., C+ER, D+S, C+EL+S, etc.).

***Synthetic Datasets***

***Dataset 1: Synthetic Process Variation Dataset (SPVD)***

To evaluate the performance of the proposed framework, particularly the Transfer Learning (TL) and semi-supervised learning (SSL) capabilities, a controlled dataset exhibiting characteristics similar to real-world wafer maps was required. Due to the proprietary nature and confidentiality constraints associated with actual semiconductor manufacturing data, we developed a procedural generation pipeline to create the large-scale Synthetic Process Variation Dataset (SPVD). This synthetic dataset allows for controlled experimentation, systematic introduction of specific defect signatures, and provides a sufficient volume of data for training and validation without compromising sensitive information. The generation process is designed to produce visually complex images with randomized features mimicking process variations and specific defect types relevant for TL/SSL benchmarking.

The SPVD was generated using a custom Python script leveraging the OpenCV (cv2) and NumPy (np) libraries. The process involves several stages designed to introduce variability and realistic features:

- **Base Pattern Generation:** An initial 400x400 pixel canvas is created with a neutral grey background. A random number (0 to 4) of concentric rings (radius 30-200 pixels, thickness 4-10 pixels, black or white) are added, with positional perturbations introduced via slightly offset grey circles to create less uniform edges.

- **Background Texturing:** A complex background texture mask is generated using blurred circular features (0-5 rings, random radii) and radial lines. This mask is heavily blurred (Gaussian kernel 65x65) and blended (cv2.addWeighted) with the ring image to create subtle background variations.

- **Colorization and Styling:** The resulting grayscale image undergoes color inversion (255 - image) and significant Gaussian blurring (kernel 25x25) to smooth features. The final colorization uses the JET colormap (cv2.applyColorMap), producing a blue-green-yellow-red appearance (See Figure [Insert Figure Number for SPVD samples]).

- **Circular Masking:** A circular mask (radius 200 pixels) is applied, setting the exterior region to black to emulate the physical wafer shape.

- **Defect Simulation (for 'bad' class):** For images designated as defective ('faulty'), simulated defects are introduced after base pattern generation but before final blurring/masking. A random number (1 to 5) of defects are rendered as short, thin radial line segments ("strokes", length 5-20 pixels) near the wafer periphery (radial distance 160-180 pixels) with randomized angles and positions (cv2.line). These strokes are slightly blurred (Gaussian kernel 3x3) to blend subtly, appearing as distinct anomaly markers (See Figure 4). Images designated as "good" skip this step.

*Dataset characteristics*

- Total Images: 200

- Resolution: 400 × 400 pixels

- Format: PNG (implied, based on standard practice)

- Classes (2 types): good: 100 images (defect-free), faulty: 100 images (containing simulated stroke defects)

- Class Balance: Perfectly balanced between 'good' and 'bad' classes.

*Key Features:* High visual variability within classes due to extensive randomization; explicitly controlled binary defect type; scalable generation script.

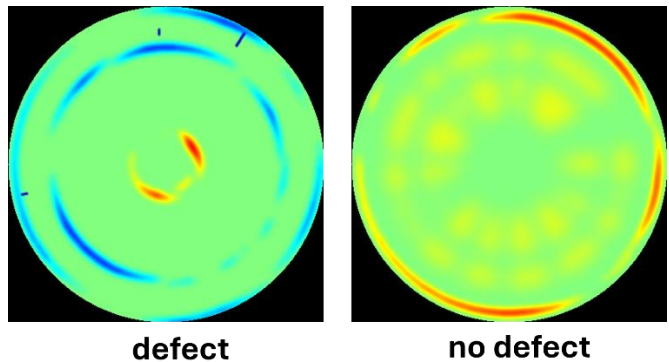

defect        no defect

Figure 4 Synthetic process variation dataset (SPVD) comprised of 200 images containing two classes: good (no defect) and faulty (defect in form of strokes) as illustrated with 100 images corresponding to each class.

Note that as a synthetic dataset, SPVD primarily models background process variations and a specific binary defect type (strokes). It may not fully represent the diversity of defect morphologies or subtle signal variations found in all real-world manufacturing scenarios.

***Dataset 2: Synthetic WM-811K Emulation Dataset (SWED)***

To further facilitate the development, testing, and benchmarking of wafer map classification approaches, particularly for multi-class spatial pattern recognition, we generated the Synthetic WM-811K Emulation Dataset (SWED). Real-world datasets like WM-811K (75) often exhibit significant class imbalance, potentially hindering the training of robust models for rare defect types. SWED provides a balanced and controlled environment, mimicking the primary spatial defect patterns found in WM-811K, thereby allowing for systematic evaluation of algorithms designed for multi-class wafer map classification.

The SWED was generated programmatically using Python, leveraging libraries including NumPy, Matplotlib (rendering), scikit-image (geometric drawing: disk, line_aa), and SciPy (ndimage for morphological operations: binary_dilation).

- **Base Wafer Creation:** A 128x128 pixel grid is initialized. A circular wafer area is defined (radius 94% of half-size). Pixels inside are designated 'non-defective' (value 1), outside are 'background' (value 0).

- **Defect Pattern Primitives:** Procedural functions generate defect signatures: localized circular clusters (disk), ring patterns (defined by radial bounds), scratches (thickened lines via line_aa and binary_dilation), and area defects (random pixel assignment).

- **Class-Specific Generation:** Specific primitive functions are invoked based on the target WM-811K class. Parameters (size, density, location, count, etc.) are randomized within predefined ranges for intra-class variability. Sparse random noise pixels are added to most non-'None' types.

- **Discrete Representation:** Pixels are assigned discrete values: 0 (background), 1 (non-defective), 2 (defective).

- **Image Rendering:** Each 128x128 discrete array is rendered as a PNG image. A fixed 3-color map ([#440154, #21918c, #fde725] mapping 0, 1, 2 to purple, teal, yellow) and nearest-neighbor interpolation are used, visually emulating common WM-811K visualizations.

*Dataset Characteristics*

- Total Images: 1,800

- Resolution: 128 × 128 pixels

- Format: PNG (Grayscale values 0, 1, 2; visualized with specific colormap)

- Classes (9 types, mimicking WM-811K):

    1. None: Minimal/no defects.

    2. Center: Central defect cluster.

    3. Donut: Ring defect offset from center/edge.

    4. Edge-Loc: Localized defects near the edge.

    5. Edge-Ring: Ring defect near the perimeter.

    6. Loc: Localized defects away from center/edge.

    7. Random: Randomly distributed defects.

    8. Scratch: Linear scratch-like defects.

    9. Near-Full: Wafer predominantly covered by defects.

- Class Balance: Perfectly balanced (200 unique samples per class).

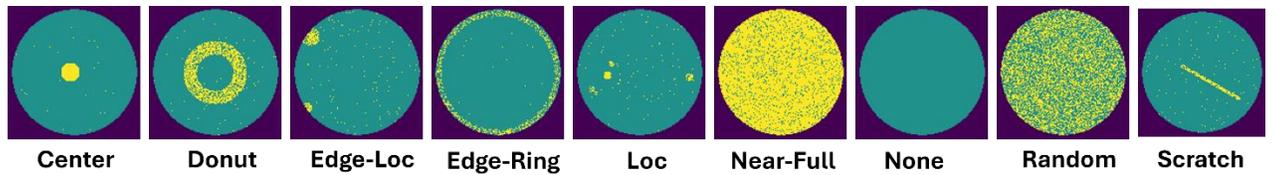

Figure 5 Synthetic WM-811K Emulation Dataset (SWED) comprised of 1800 images belonging to 9 classes as illustrated with 200 images corresponding to each class.

Note that while visually emulating WM-811K patterns, SWED is synthetic. It may not capture the full complexity of real defect morphologies, subtle process shifts affecting background signals, or specific instrumentation noise profiles. The generation relies on predefined geometric and stochastic rules.

**Results**

In this section, we present the results obtained by applying the integrated Deep Topological Data Analysis (Deep TDA), Self-Supervised Learning (SSL), and Transfer Learning (TL) framework to the diverse datasets described previously (WM811K, Mixed WM38, SPVD, SWED). The experiments were designed to evaluate the framework's capabilities under different conditions, including clustering complex real-world data using the core SSL+Deep TDA pipeline and assessing the effectiveness of zero-shot transfer learning with a pre-trained foundational model on synthetic datasets. Table 1 provides a summary overview of these experimental use cases, outlining the dataset, framework configuration, primary objective, and key outcomes for each analysis. Detailed findings for each use case are presented in the subsequent subsections.

Table 1 Overview of the different datasets and framework configurations used to evaluate the proposed integrated approach. Each row represents a distinct experimental setup designed to test specific capabilities, such as handling real-world imbalance (Use Case 1), multi-label complexity (Use Case 2), or zero-shot transfer learning performance on controlled synthetic data (Use Cases 3 & 4).

| Use case | Dataset used | Framework Configuration | Primary Objective | Key Outcome/Demonstration | Relevant Figures |
|---|---|---|---|---|---|
| 1 | WM811K (Open source) | Core SSL + Deep TDA Pipeline | Validate baseline unsupervised clustering on real-world, imbalanced single-defect data. | Effective separation of major defect categories without labels; sensitivity to rare patterns; handling of class imbalance. | 6-11 |

| 2 | Mixed WM38 (Open source) | Core SSL + Deep TDA Pipeline | Assess framework's ability to cluster complex data with single and mixed defect types. | Successful segmentation reflecting single and multi-label patterns; sensitivity to combined defect signatures; handling high complexity. | 12-17 |
| 3 | SPVD (Synthetic) | Pre-trained Model (SSL+Deep TDA+**TL**) | Evaluate zero-shot transfer learning for binary classification (good/faulty) with background process variation. | Successful zero-shot separation of good/faulty classes; further separation of 'faulty' based on background, showing sensitivity to process variation via TL. | 4,18-19 |
| 4 | SWED (Synthetic) | Pre-trained Model (SSL+Deep TDA+**TL**) | Validate zero-shot transfer learning on balanced, controlled multi-class data mimicking WM-811K patterns. | Correct grouping of distinct vs. visually similar synthetic defect classes (e.g., pure Donut/Edge-Ring clusters, merged Loc/Edge-Loc); validation of pre-trained model's representational quality for WM-811K-like patterns. | 5,20-22 |

### *Image Clustering of WM811K Dataset*

The proposed integrated framework of Deep TDA and SSL as described in Figure 1 is implemented on the WM811K dataset comprising of the 8 failure categories. The objective here is to separate the images based on the failure categories. This analysis has been conducted on 25, 527 images containing patterns specific to the mentioned 8 failure categories: Center, Donut, Edge-Loc, Edge-Ring, Loc, Near-Full, Random and Scratch as shown in Figure 6. The dataset is imbalanced with majority of the patterns represented by the Edge region defects which are in general difficult to characterize. The image processing required here is minimal, all images are resized to have the same size. The hyperparameters specified for the TDA grid search and training the SSL learning framework in the DataRefiner platform are specified in Table 2. The resulting TDA map demonstrating the separation of these 25k+ images based on their characteristic patterns is shown in Figure 7. The TDA map is resultant of the TDA grid search and tunning of the SSL network. The TDA map with the lowest score is chosen for analysis. The TDA map in Figure 7 is chosen based on the lowest score post TDA grid search. The resulting TDA parameters for this map are Beta of 3.5 and Metric is Euclidean. The no of clusters (or networks, used interchangeably) is 4 where the largest cluster contains 69% of the total images. The resulting distribution of the failure categories per cluster is shown in Figure 8 where the cluster with largest image count is shown to have the Edge-Loc and Center as major failure categories. The second most populated cluster comprises of Edge-Ring as the most representative category in that cluster. Rest of the clusters are dominated by the Edge-Ring patterns which is not surprising considering this category to be the most represented failure mode in the dataset considered. There are few images that do not get assigned to any structure thus, get assigned to the Noise cluster. In Figures 9-11 individual segments of the TDA map as illustrated in Figure 7 are analyzed for each failure category. The distribution of the failure categories in each of the extracted segments shown indicates the uniqueness and overlapping characteristics of their patterns. As observed in these individual segments, the core of each of the network comprises of the patterns that are represented by majority of the images of a particular failure type in that network ex. 'Edge-Loc' Figure 10.f whereas the periphery of the network represents the most unique pattern ex. 'Near-Full' Figure 9.c.

Table 2 Hyperparameters settings in DataRefiner© platform to execute image clustering on WM811K dataset comprising of 25,527 images with 8 failure categories.

| TDA Grid search | Size | Model complexity | Augmentations | No. of epochs | Batch size | Learning Rate |
|---|---|---|---|---|---|---|
| Beta = 3.5, 10.0, 20.0 Metric = Euclidean, Cosine | 35x35 | Restnet18 | Horizontal flip, Vertical flip, Rotation (0,45 degrees) | 600 | 512 | 0.12 |

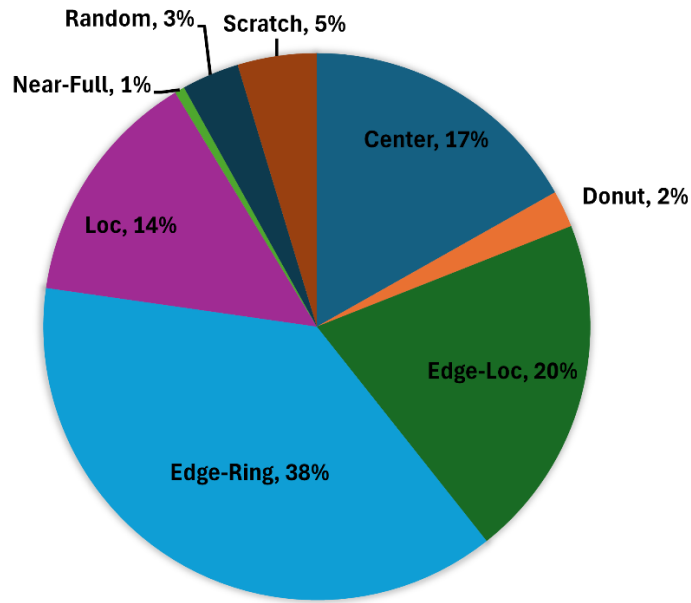

Figure 6 Distribution of Failure Categories in the WM811K Dataset: This pie chart illustrates the proportion of 25,527 images across eight distinct failure categories. The 'Edge-Ring' category accounts for the largest share at 38%, followed by 'Edge-Loc' at 20%, and 'Center' at 17%. Other categories include 'Loc' (14%), 'Scratch' (5%), 'Random' (3%), 'Donut' (2%), and 'Near-Full' (1%).

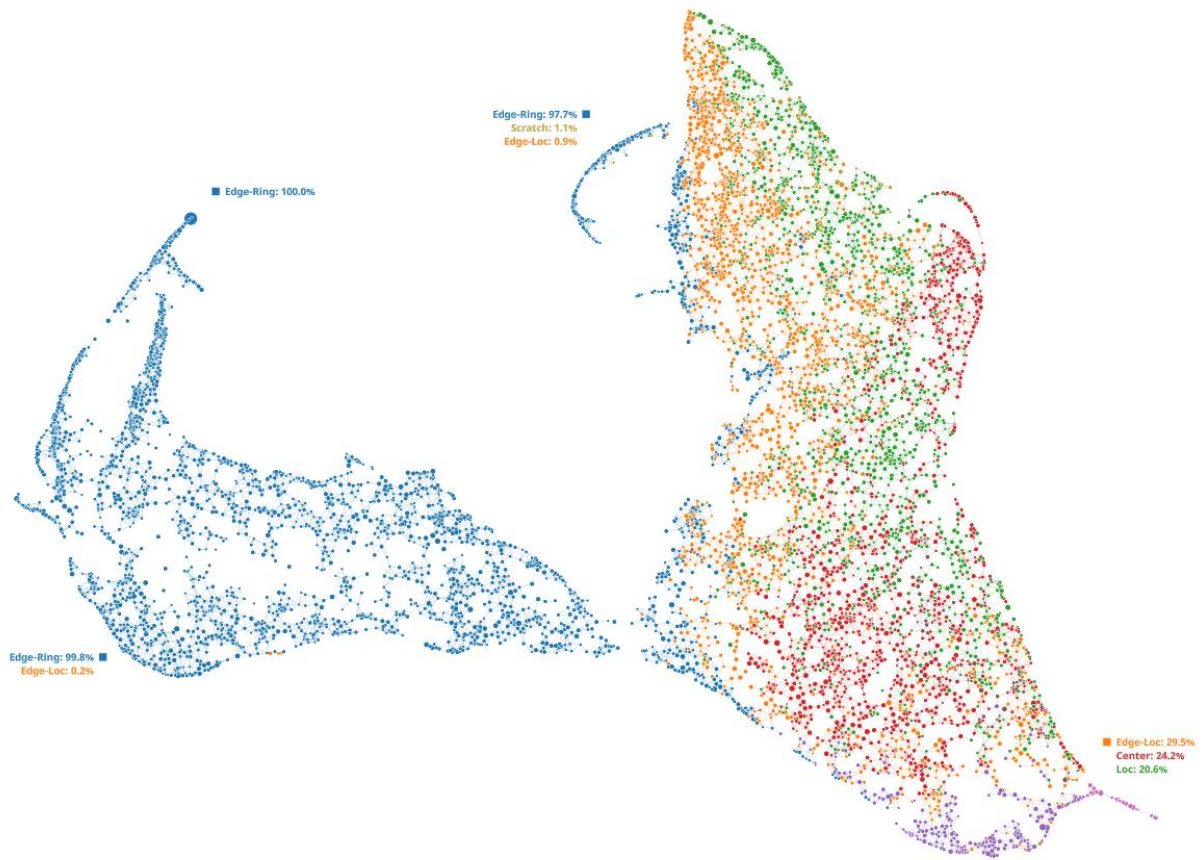

Figure 7 TDA map illustrating the segmentation of 25, 527 wafer images of the WM811K dataset into 4 clusters (or networks) shown. Note that the nodes in the network represent the images that have similar characteristics and edges connecting the nodes are indicative of images that share overlapping characteristics. The colors are used to represent the different clusters. The labels of the defect categories are used for illustrating the distribution of the various defect categories in these networks. Labels of these images were not part of the learning framework. The percentage distribution of each failure category associated with each network is displayed. For ex, the cluster 'Edge-Ring' is composed 100% of the images corresponding to the failure category 'Edge-Ring'.

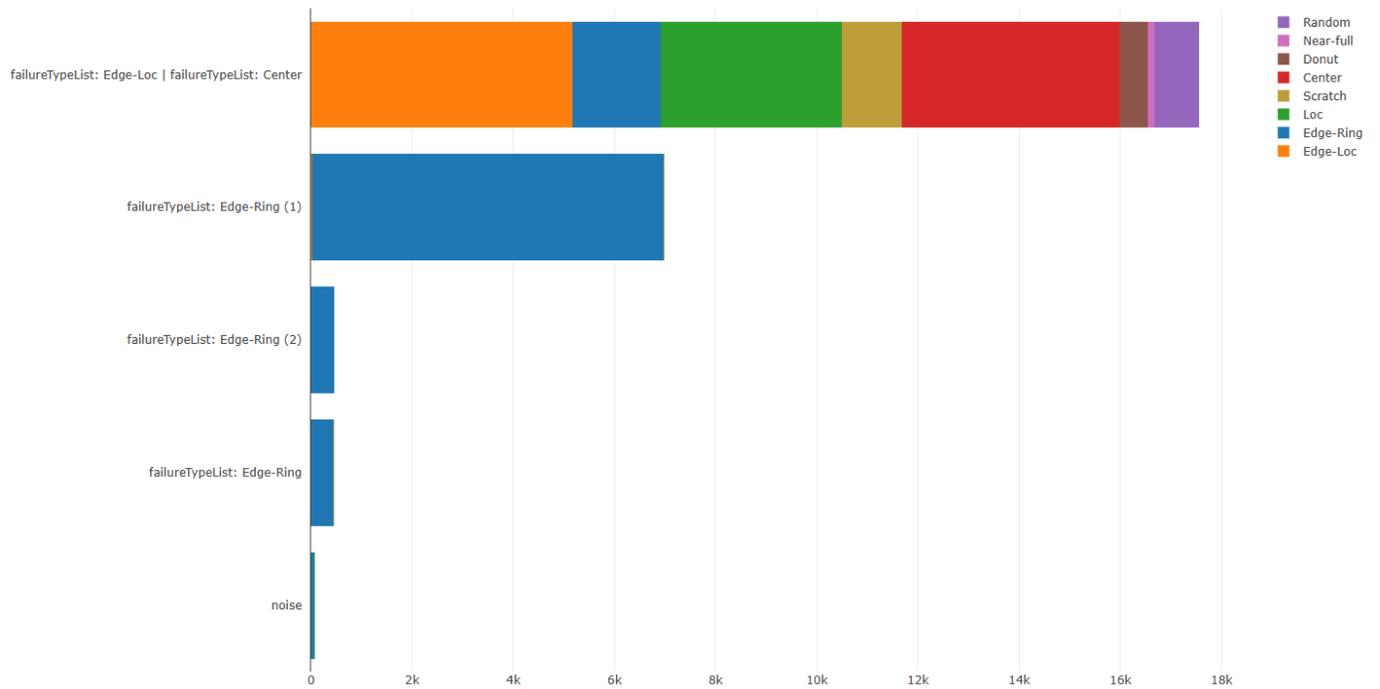

Figure 8 Histogram illustrating the distribution of each failure category across clusters. The X-axis represents the count of images, while the Y-axis displays the cluster names, which denote the most representative category within each cluster. For instance, in the largest cluster labeled 'failureTypeList: Edge-Loc| failureTypeList: Center', the predominant categories are 'Edge-Loc' and 'Center'.

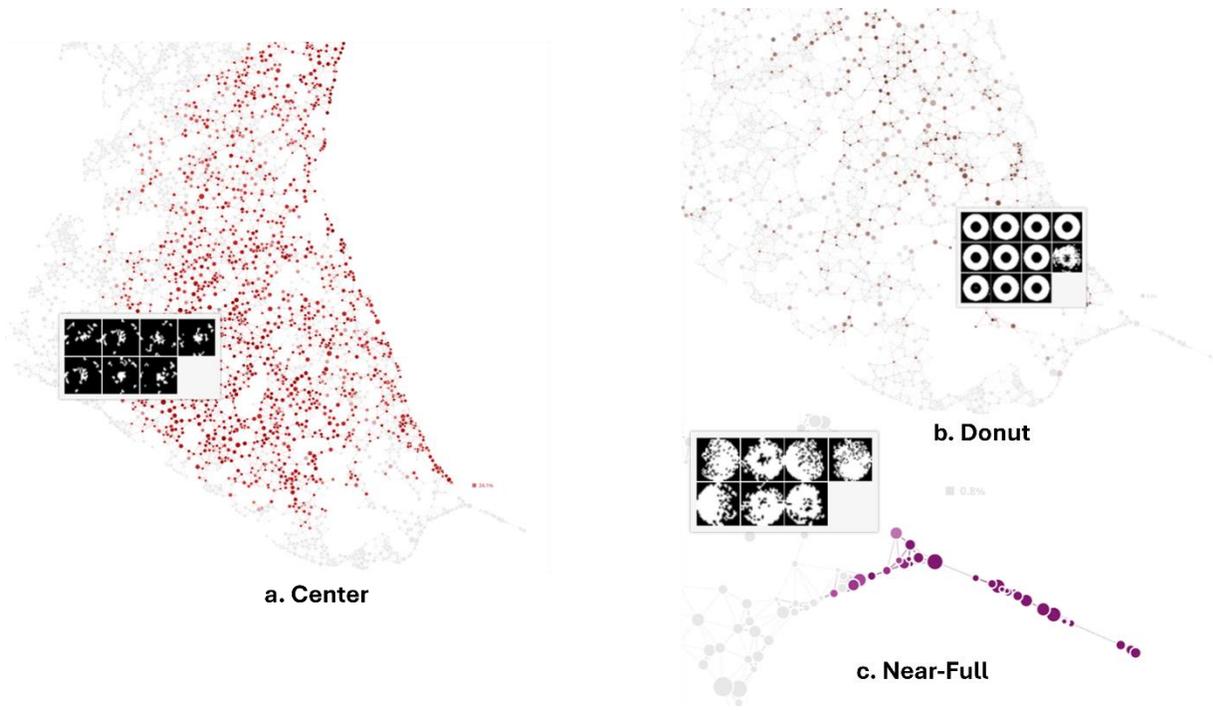

Figure 9 Individual segments per failure category analyzed from the TDA map shown in Figure 7. Categories represented are: a. Center, b. Donut, and c. Near-Full. It is quite impressive to see TDA aggregating the patterns corresponding to the 'Near-Full' category appear in the tail region of the network indicating their uniqueness compared to the rest of the failure patterns in that network.

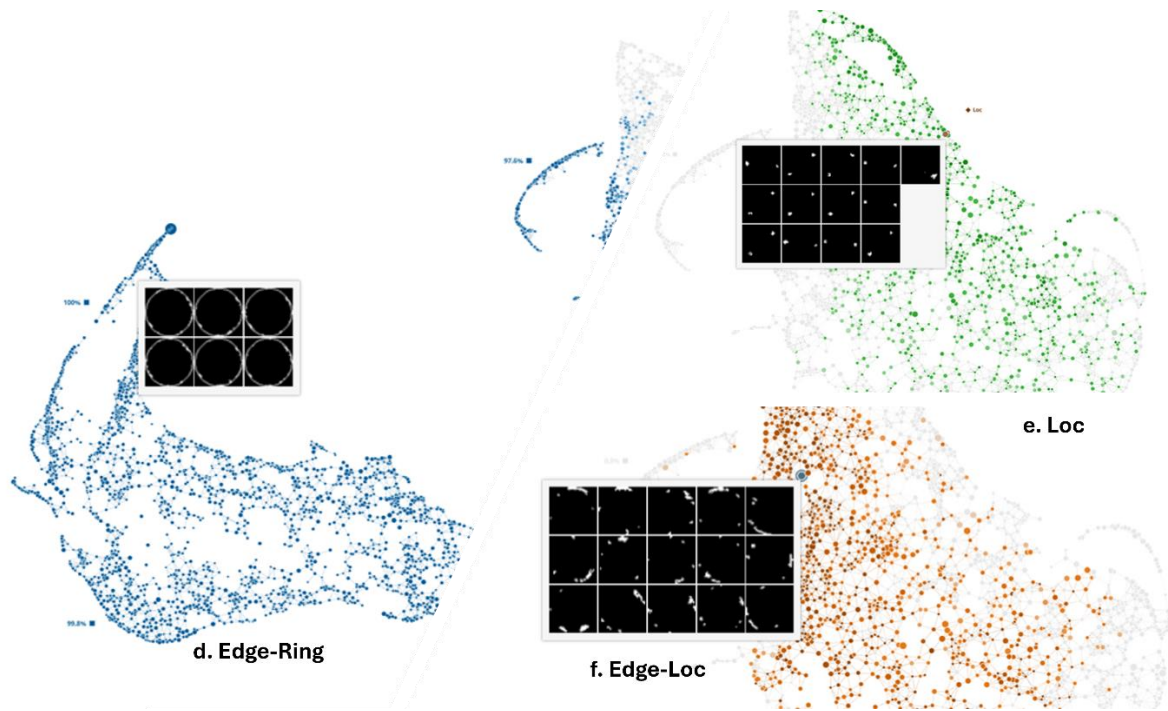

Figure 10 Continuation of the individual segment analysis per failure category from the TDA map depicted in Figure 7. The categories highlighted here include: d. Edge-Ring, e. Loc, and f. Edge-Loc. The patterns associated with each failure category are displayed to illustrate their overlapping characteristics, particularly the 'Edge-Loc' and 'Loc' categories, which frequently appear in the same cluster labeled 'failureTypeList: Edge-Loc| failureTypeList: Center' as shown in Figure 7. The 'Edge-Ring' pattern, being the most prevalent, results in the majority of the 9,681 images being segmented into three out of the four clusters depicted.

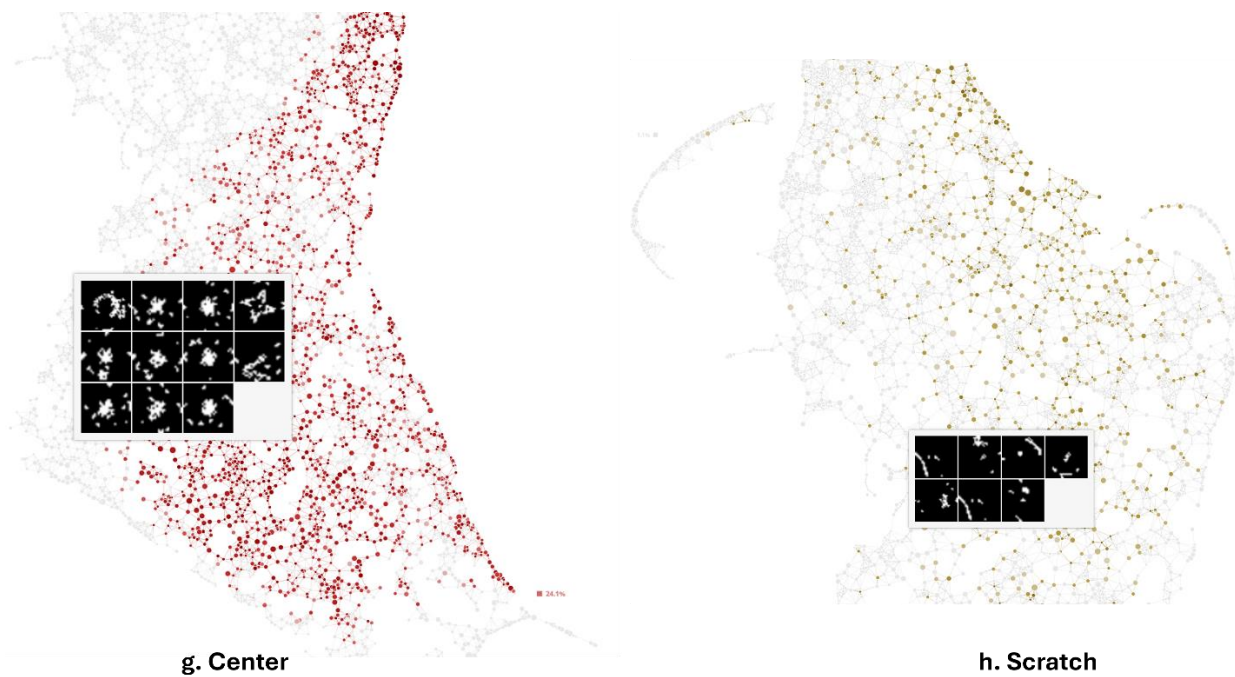

Figure 11 The final two failure categories, g. Center and h. Scratch, are depicted in these segments extracted from the TDA map in Figure 7. The 'Scratch' failure is more dispersed throughout the network compared to the 'Center', owing to its noisy pattern and similarity to the 'Loc' pattern.

### *Image Clustering of Mixed 38WM Dataset*

The integrated framework combining Deep TDA and SSL, as outlined in Figure 1, is further applied to the Mixed 38WM open-source dataset, as described in the Datasets section. Image clustering is carried out on 38,015 wafer maps composed of 8 categories of single defect types, 1 normal type and 3 kinds of mixed defect types synthesized by combining single defect type resulting in 29 mixed-type defects (76-77) as shown in Figure 12. The image processing required here is minimal, all images are resized to have the same size. The hyperparameters specified for the TDA grid search and training the SSL learning framework in the DataRefiner platform are specified in Table 3 . The resulting TDA map demonstrating the separation of these 35k+ images based on their characteristic patterns is shown in *Figure 14*. The TDA map is resultant of the TDA grid search and tunning of the SSL network. The TDA map with the lowest score is chosen for analysis. The TDA map in *Figure 14* is chosen based on the lowest score post TDA grid search. The resulting TDA parameters for this map are Beta of 10.0 and Metric is Euclidean. The number of clusters (or networks, used interchangeably) is 31 where the largest cluster contains 16% of the total images.  The resulting distribution of the failure categories per cluster is shown in Figure *8* Figure 15 where the cluster with largest image count is shown to have the Edge-Loc, Edge_Ring and Scratch as major single type failure categories. The resulting mixed type categories Edge_Loc+Scratch and

Edge_Ring+ Scratch are thus, represented in that cluster. Rest of the clusters follow the same trend in terms of the representative defect type categories. In Figure 16 and Figure 17 the image segmentation results for single-type defects (a. C7:Near-Full and b. C9:Random) and mixed-type defects (C28:Donut+Edge-Loc+Loc and C30: Donut+Edge-Ring+Loc) are shown. The image segmentation results for the single-type defects (Figure 16) reveal the percentage distribution of defect modes within the network, emphasizing their shared characteristics that place them in the same neighborhood, while the distinctiveness of the 'Random' pattern is highlighted by its position in the network's 'tail' region. Additionally, the segmentation results for mixed-type defects (Figure 17) are compared across two distinct networks, showing not only a variation in image counts but also highlighting pattern distinctiveness, suggesting that these defects may arise from different root causes.

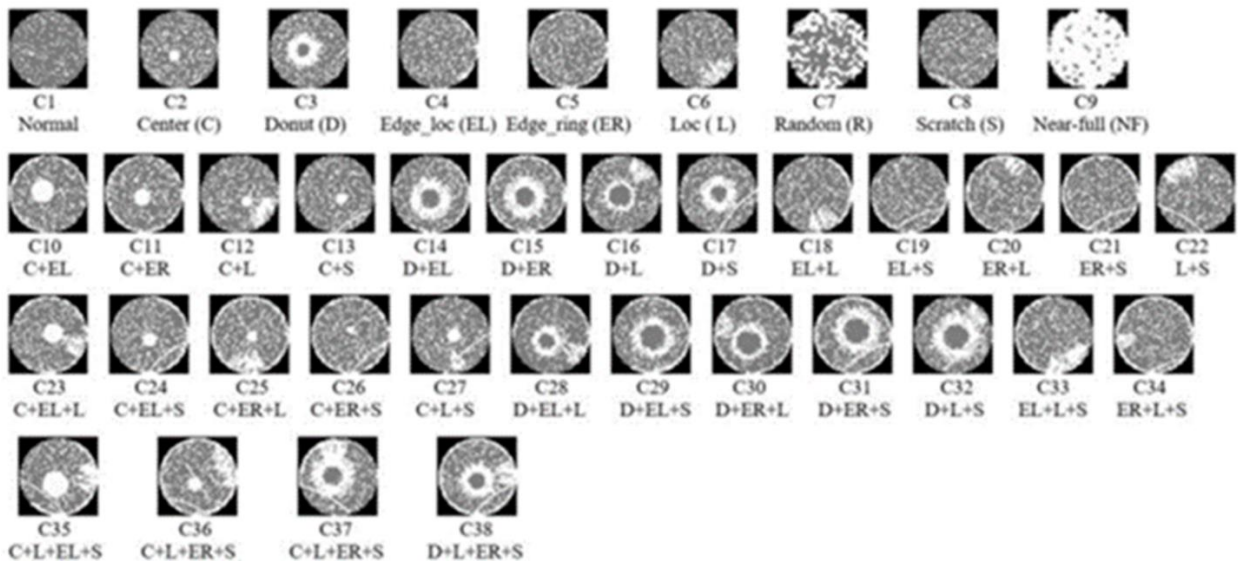

Figure 12 Sample images from the Mixed 38WM dataset showcasing the diverse spatial defect patterns found on wafer maps. The top row shows the single defect classes (C1-C9), while the lower rows show the complexity of multi-label defects (C10-C38) where multiple patterns occur simultaneously.

Table 3 Hyperparameters settings in DataRefiner© platform to execute image clustering on Mixed 38 WM dataset comprising of 38,015 images with 1 normal and 37 failure categories.

| TDA Grid search | Size | Model complexity | Augmentations | No. of epochs | Batch size | Learning Rate |
|---|---|---|---|---|---|---|
| Beta = 3.5, 10.0 Metric = Euclidean | 32x32 | Restnet18 | Horizontal flip, Vertical flip, crop, Rotation (0,180 degrees) | 1000 | 256 | 0.12 |

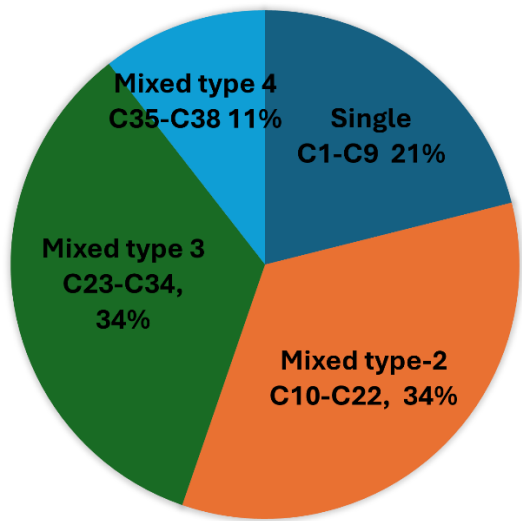

Figure 13 Composition of the analyzed wafer Mixed 38WM dataset, highlighting the prevalence of multi-label defects. Single defect patterns (C1-C9) constitute 21% of the data, while mixed types involving two (34%), three (34%), or four (11%) simultaneous defect patterns make up the majority (79%).

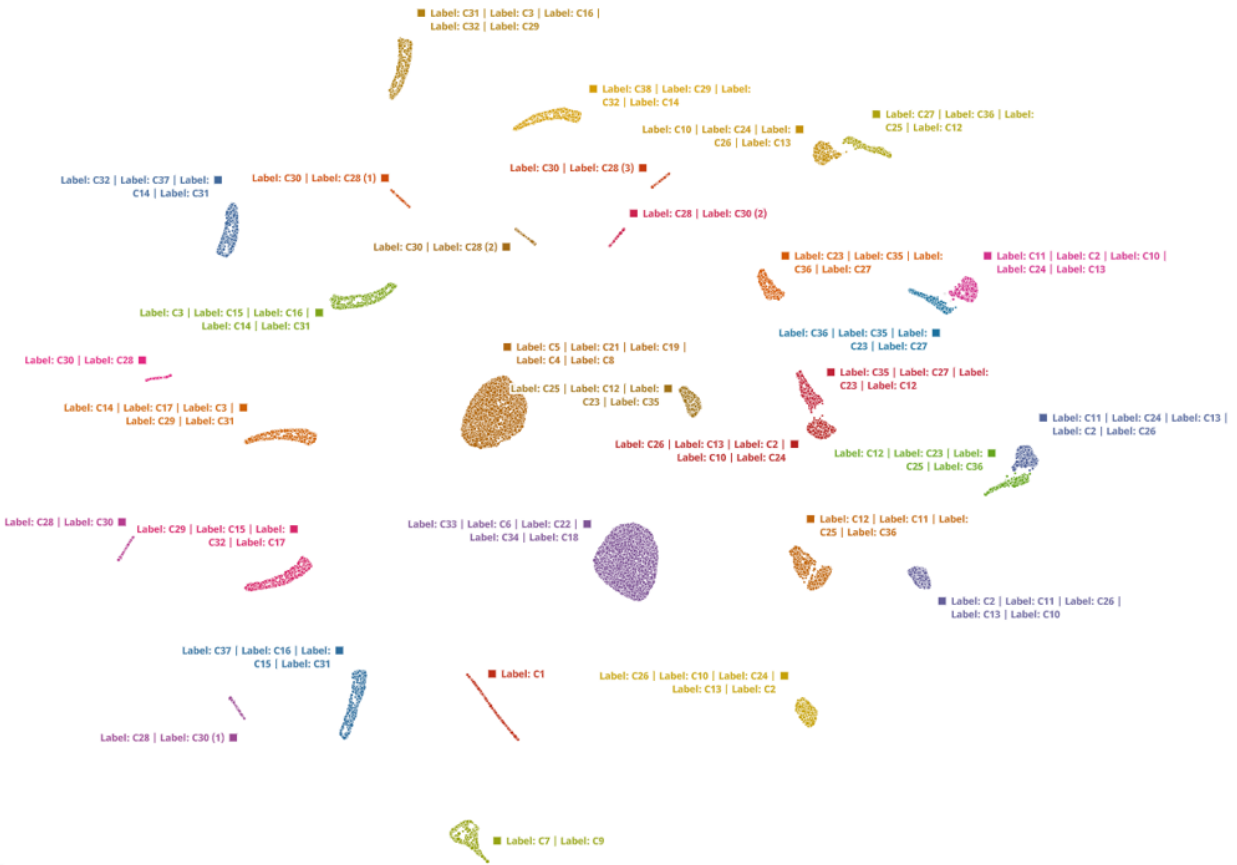

Figure 14 TDA map illustrating the segmentation of 35k+ wafer images of the Mixed WM38 dataset into 31 clusters(/networks) shown. Different colors are used to denote various clusters. The labels of defect categories serve to illustrate the distribution of these categories across the networks. Labels of these images were not included in the learning framework. The percentage distribution of each failure category within each network is shown where the most representative category in that network is indicated in the name of the cluster for ex.' Label: C1' (normal) category comprising entirely of normal images where as 'Label: C7 | Label: C9' composed of categories C7 (Random) and C9 (Near-Full).

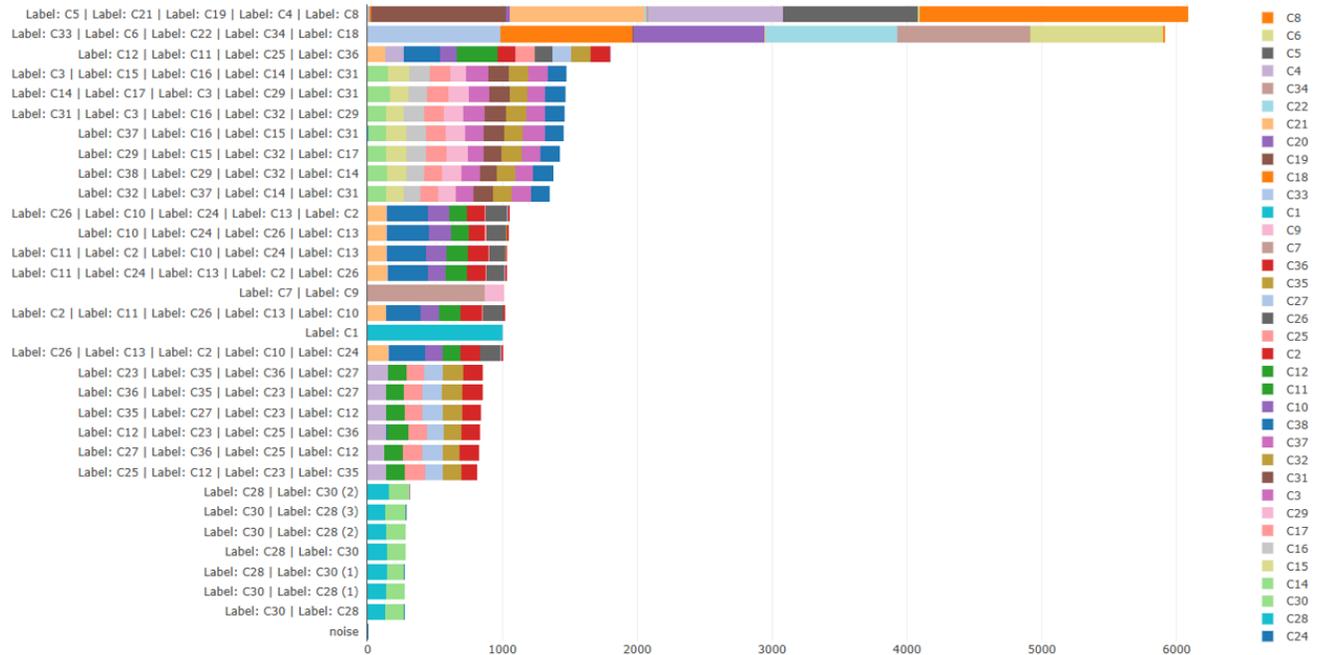

Figure 15 Histogram illustrating the distribution of each failure category across clusters. The X-axis represents the count of images, while the Y-axis displays the cluster names, which denote the most representative category within each cluster. For instance, in the largest cluster labeled 'Label: C5| Label: C21| Label: C19| Label: C4| Label: C8', the predominant single type defect categories are 'C4: Edge_Loc', 'C5: Edge_Ring', and 'C8: Scratch' which then combine to form the represented mixed defect categories 'C19:Edge_Loc+ Scratch' and 'C21: Edge_Ring+Scratch'.

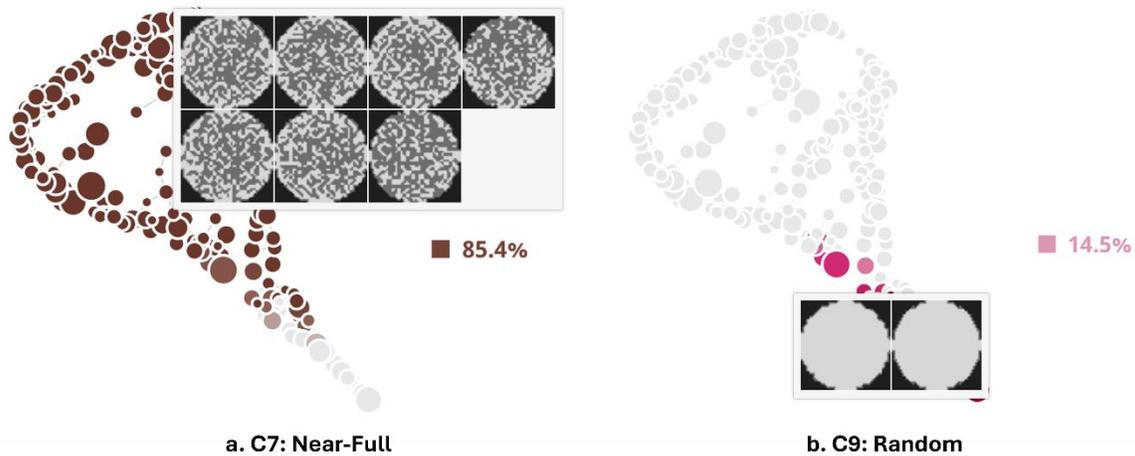

a. C7: Near-Full       b. C9: Random

Figure 16 The image segmentation results for single-type defects are presented, illustrating the percentage distribution of C7 (Near-Full) and C9 (Random) defect modes in the shown network. It is important to note that the patterns for both defect types share characteristics,

causing them to appear in the same neighborhood. The uniqueness of the 'Random' pattern is emphasized by its position in the 'tail' region of the network.

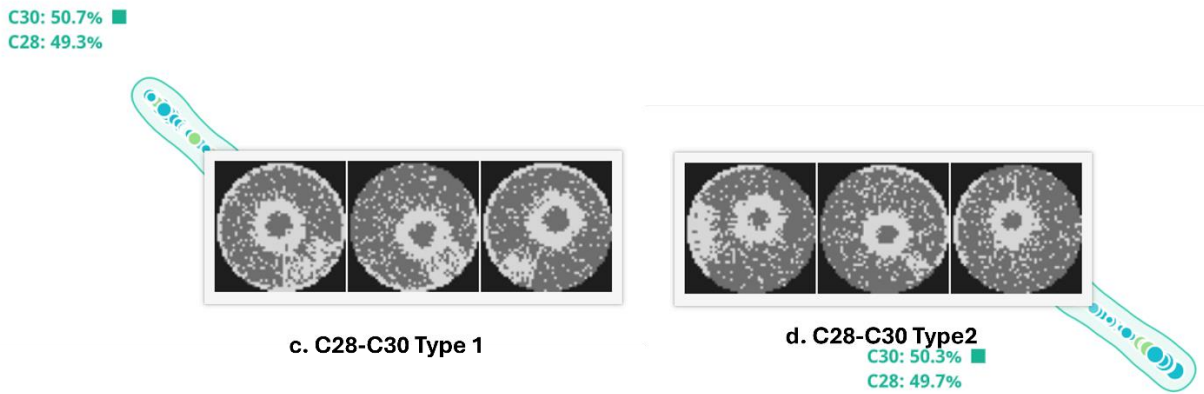

Figure 17 The image segmentation results for mixed-type defects, involving three combined defects, are displayed. The percentage distribution of C28 (Donut+Edge-Loc+Loc) and C30 (Donut+Edge-Ring+Loc) defect modes across two distinct networks are shown and compared. Apart from the obvious variation in count of images between the two networks, the segmentation highlights the pattern distinctiveness, indicating that the defects may stem from different root causes.

*Image Clustering of Synthetic Process Variation Dataset (SPVD)*

Building upon the robust Self-Supervised Learning (SSL) framework integrated with Deep Topological Data Analysis (Deep TDA) as previously detailed in Figure 1, we now present the results achieved by incorporating Transfer Learning (TL), applied to the Synthetic Process Variation Dataset (SPVD), as described in the Datasets section. The dataset consists of 200 images. It includes two classes, each with 100 unique samples, mimicking defects introduced during varying process and a specific binary defect type (strokes): Good (no defects) and Faulty (defect) as shown in Figure 4. The dataset is perfectly balanced.

The hyperparameters specified for the TDA grid search and implemented for the pre-trained large image model in the DataRefiner platform are specified in Table 4. The resulting TDA map with the lowest score corresponds to Beta=3.5 and Metric=Euclidean is shown in Figure 18. The number of clusters (or networks) is 20 with largest cluster comprising of 11% of the images where each cluster represents either the 'good' or 'faulty' class wafer images. The distinction between these clusters arises due to the inherently different backgrounds and 'stroke' patterns as shown in Figure 19.

Table 4 Hyperparameters settings in DataRefiner© platform to execute image clustering using the pre-trained image model on the SPVD, synthetic wafer map dataset comprising of 200 images with 1 'good' and 1 'faulty' categories.

| TDA Grid search | Size | Model complexity | Augmentations | No. of epochs | Batch size | Learning Rate |
|---|---|---|---|---|---|---|
| Beta = 1.5, 3.5 Metric = Euclidean, Cosine | 384x384 | Restnet18 | crop, Rotation (0,45 degrees) | 100 | 256 | 0.12 |

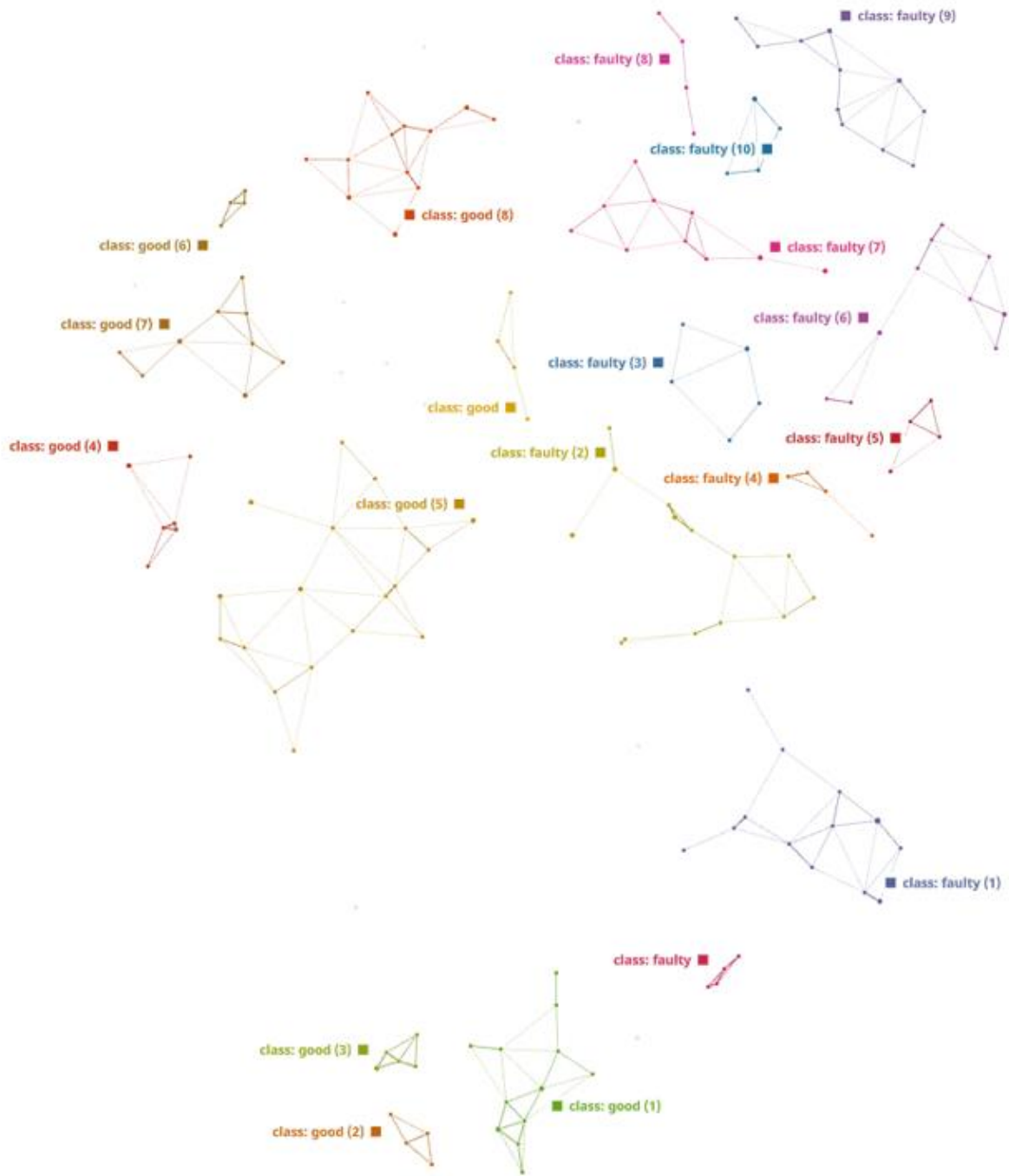

Figure 18 TDA map illustrating the segmentation of 200 wafer images of the synthetic dataset SPVD into 20 clusters (or networks) shown. Note that the colors are used to represent the different clusters, and their names correspond to the most represented category in that cluster. The labels of the defect categories are used for illustrating the distribution of the various defect categories in these networks. For ex. the cluster 'faulty' is composed 100% of the images corresponding to the failure category 'faulty like the cluster 'good' which is comprised entirely of images with no defect.

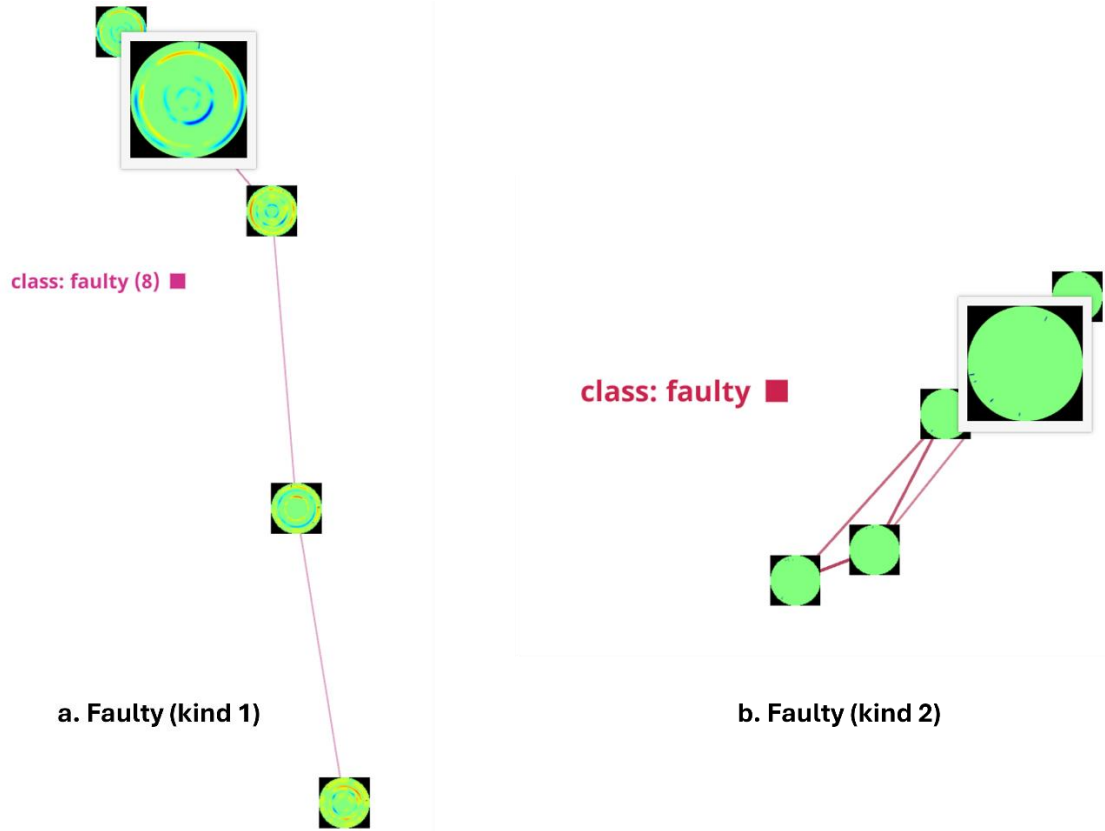

Figure 19 Illustrating the segments of the TDA map with overlayed wafer map images derived from the synthetic dataset SPVD which reveals that the wafer images classified as 'faulty' get separated into two distinct clusters shown (a. Faulty (kind1) and b. Faulty (kind2)), based on the defect patterns and their underlying backgrounds. In real world scenario, these patterns could suggest varying process variations leading to the observed defects. Additionally, the notable differences in the image backgrounds imply that they might have been processed using different tools or during different process steps.

*Image Clustering of Synthetic WM-811K Emulation Dataset (SWED)*

Next we apply the DTDA+SSL+TL framework, to the Synthetic WM-811K Emulation Dataset (SWED), as described in the Datasets section. The dataset consists of 1,800 images. It includes nine classes, each with 200 unique samples, mimicking the WM-811K dataset: None (minimal/no defects), Center (central defect cluster), Donut (ring defect offset from center/edge), Edge-Loc (localized defects near the edge), Edge-Ring (ring defect near the perimeter), Loc (localized defects away from center/edge), Random (randomly distributed defects), Scratch (linear scratch-like defects), and Near-Full (wafer predominantly covered by defects) as shown in Figure 5. The hyperparameters specified for the TDA grid search and implemented for the pre-trained large image model in the DataRefiner platform are specified in Table 5. The number of clusters (or networks) is 8 with largest cluster comprising of 22% of the images where each cluster corresponds to a given class except the 'Edge_Loc' and

'Loc' defects which share one cluster Figure 20. Occasionally, the 'Edge_Loc' and 'Loc' patterns share overlapping characteristics with certain patterns such as 'Center', and 'None' as seen in Figure 21 and Figure 22.

Table 5 Hyperparameters settings in DataRefiner© platform to execute image clustering using the pre-trained image model on the SWED, synthetic wafer map dataset comprising of 1800 images with 1 normal and 8 defect categories.

| TDA Grid search | Size | Model complexity | Augmentations | No. of epochs | Batch size | Learning Rate |
|---|---|---|---|---|---|---|
| Beta = 3.5, 10.0,20.0 Metric = Euclidean, Cosine | 384x384 | Restnet18 | crop, Rotation (0,45 degrees) | 100 | 256 | 0.12 |

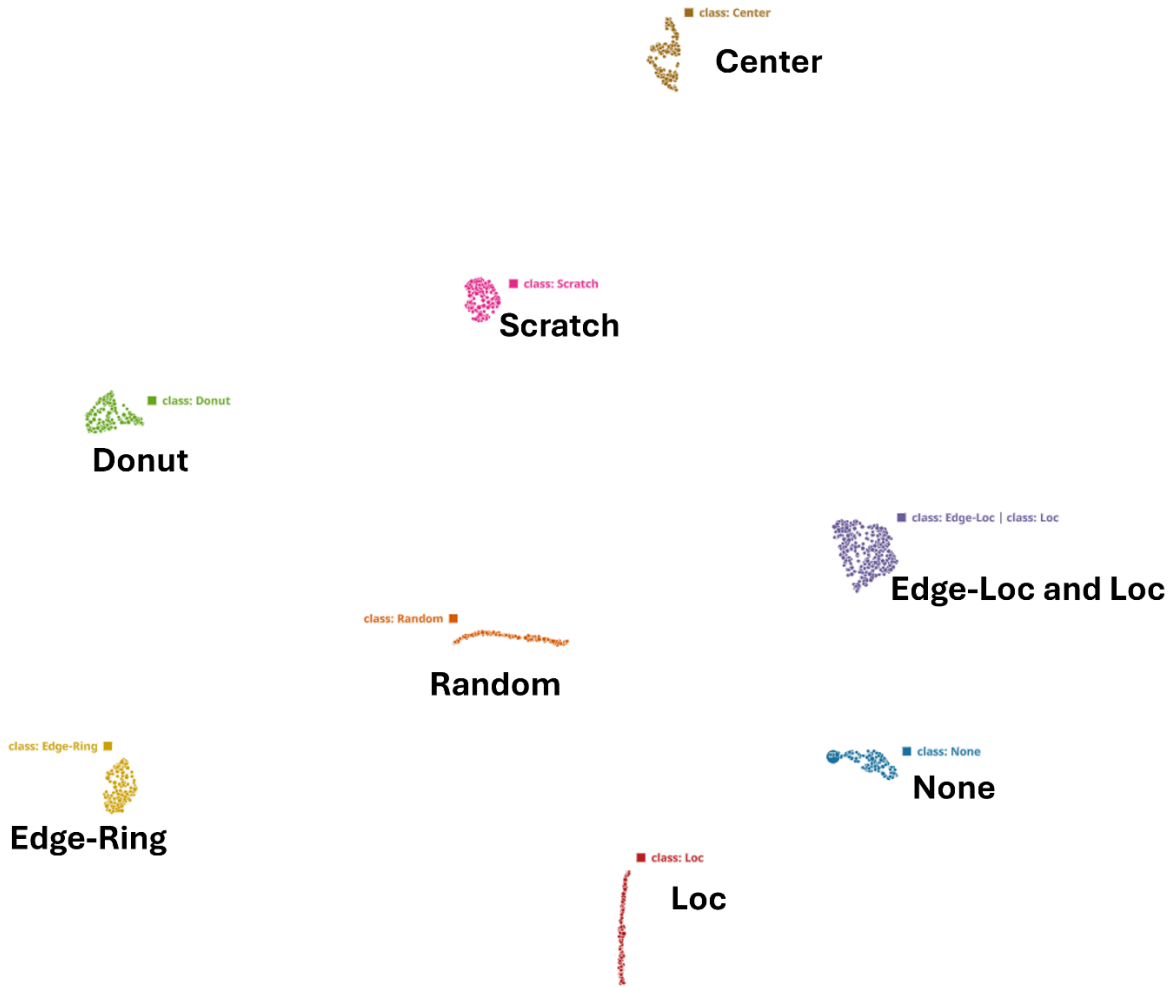

Figure 20 TDA map illustrating the segmentation of 1800 wafer images of the synthetic dataset SWED into 8 clusters (or networks) shown. Note that the colors are used to represent the different clusters, and their names correspond to the most represented category in that cluster. The labels of the defect categories are used for illustrating the distribution of the various defect categories in these networks. For ex. the cluster 'Edge-Ring' is composed 100% of the images corresponding to the failure category 'class: Edge-Ring' similar to cluster 'class: Donut' which is comprised entirely of images with 'Donut' defect type.

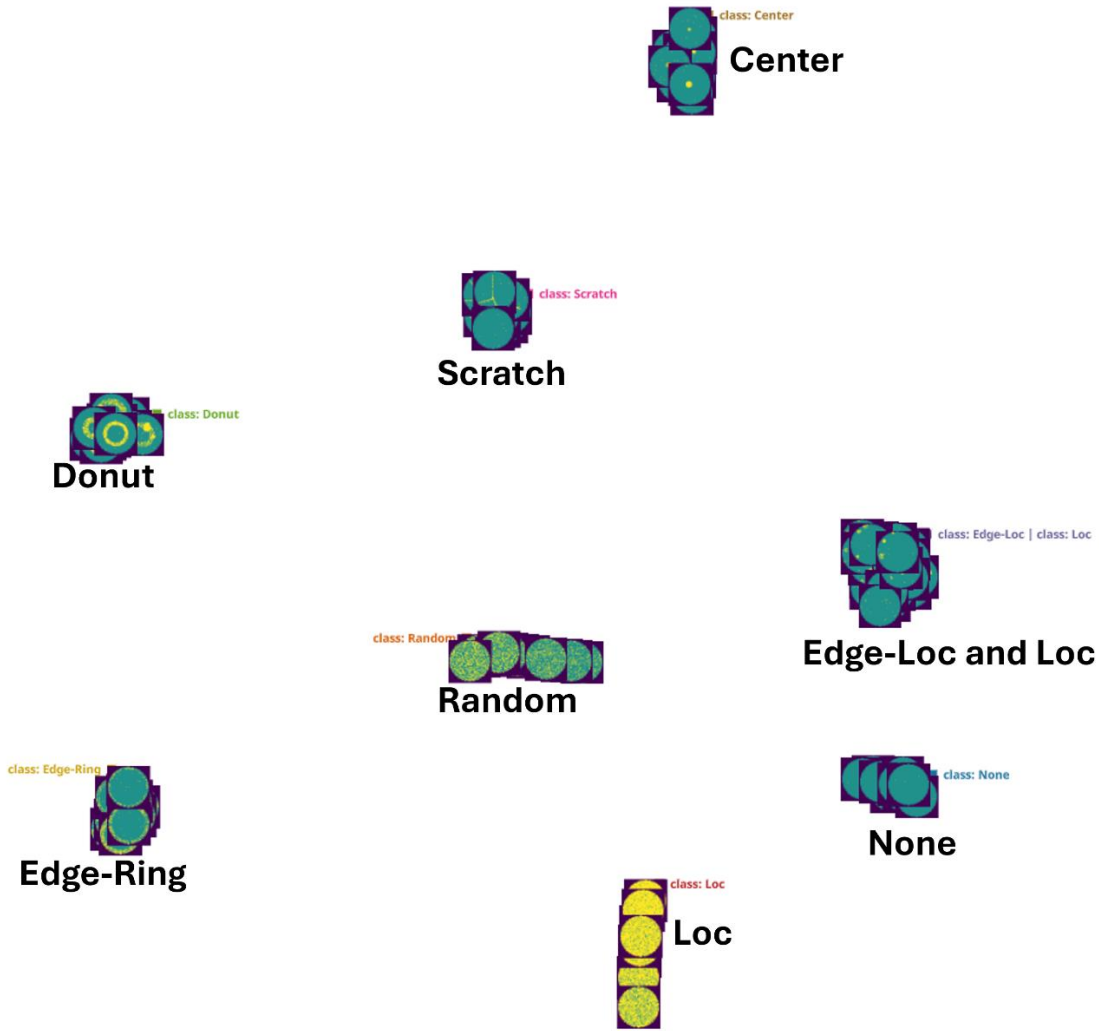

Figure 21  TDA map of synthetically generated wafer map image dataset SWED with the corresponding wafer map images overlayed illustrating the separation of 1800 wafer images into 8 clusters (or networks). Notice the separation into distinct categories based on the defect's characteristic pattern. 'Edge-Loc' and 'Loc' have overlapping characteristics, thus forming one cluster as displayed 'class: Edge_Loc | class: Loc'.

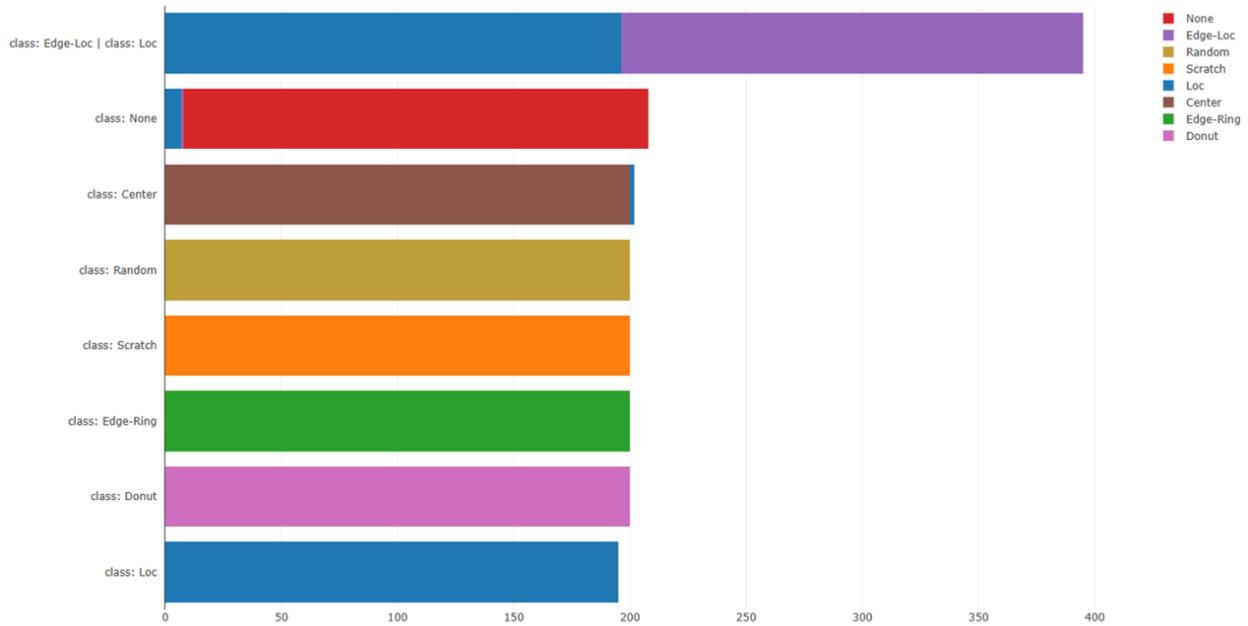

Figure 22 Histogram illustrating the distribution of each failure category across clusters for the SWED dataset. The X-axis represents the count of images, while the Y-axis displays the cluster names, which denote the most representative category within each cluster. The largest cluster here is the "class: Edge_Loc | class: Loc' which contains images from the classes 'Edge_Loc' and 'Loc' corresponding to their shared pattern characteristics. These 2 categories also appear in clusters 'class:None' (both, 'None':96.2%, 'Loc': 3.4%, 'Edge_Loc': 0.5%) and 'class:Center' ('Loc' only,'Center':99.0%, 'Loc':1.0% ) due to certain patterns which resemble the majority of the pattern type represented in the given cluster.

**Discussion**

This study introduced and validated an advanced unsupervised clustering framework integrating Deep Topological Data Analysis (Deep TDA), Self-Supervised Learning (SSL), and Transfer Learning (TL) for semiconductor image analytics. The results obtained across four distinct datasets—two real-world open-source (WM811K, Mixed WM38) and two synthetic (SPVD, SWED)—demonstrate the effectiveness and versatility of this synergistic approach.

The application of the SSL+ Deep TDA pipeline (without TL) to the WM811K dataset successfully segregated images into clusters that strongly correlated with the known, albeit highly imbalanced, failure categories (Figure 7, Figure 8). Notably, the framework achieved this separation without utilizing any labels during training, relying solely on learning discriminative representations through contrastive learning augmented by topological priors. The TDA map revealed distinct clusters dominated by major categories like 'Edge-Ring' and 'Edge-Loc'/'Center', while also isolating unique, low-frequency patterns like 'Near-Full' into specific network regions (Figure 9c), showcasing the sensitivity to both common and rare structures. The clustering of prevalent edge defects ('Edge-Ring', 'Edge-Loc') across multiple networks (Figure 10d, 10f) reflects the inherent visual similarity and variability within these classes in the real-world dataset, a challenge effectively navigated by the unsupervised approach.

Further validation on the complex Mixed WM38 dataset underscored the framework's capability to handle multi-label defect scenarios (Figure 14). The generation of 31 distinct clusters from data containing single and mixed defect types (up to four concurrent patterns) indicates a high degree of sensitivity in the learned representations. Clusters were often dominated by specific single patterns and their corresponding mixed-type combinations (Figure 15), demonstrating that the integrated TDA and SSL features capture characteristics relevant to both individual and combined defect signatures. The separation of different mixed-type defects (e.g., C28 vs. C30, Figure 17) into distinct networks, despite sharing some constituent patterns, suggests the framework may be capturing subtle differentiating features potentially linked to distinct underlying process issues or root causes, a promising avenue for diagnostic applications. The isolation of patterns like 'Random' into specific network 'tail' regions (Figure 16) further corroborates the ability to distinguish unique spatial signatures.

The experiments utilizing the pre-trained foundational model on the synthetic datasets (SPVD and SWED) successfully demonstrated the power of the Transfer Learning (TL) component. On SPVD, the framework achieved effective zero-shot clustering, separating 'good' images from 'faulty' images containing subtle 'stroke' defects (Figure 18). Significantly, the clustering further subdivided the 'faulty' class based on underlying background variations (Figure 19), indicating that the pre-trained model, enhanced by extensive exposure to diverse visual and topological features (including those captured by Deep TDA during its training), retains sensitivity to nuanced process signatures beyond just the explicit defect type. This highlights the potential for using the pre-trained model for exploratory analysis and identifying unexpected process drifts.

The application to SWED, designed to emulate WM811K in a controlled, balanced setting, confirmed the framework's ability to reconstruct expected relationships between defect classes using the pre-trained model (Figure 20). Distinct patterns like 'Donut' and 'Edge-Ring' formed pure clusters, while visually similar patterns 'Edge-Loc' and 'Loc' were correctly grouped together (Figure 21, Figure 22), mirroring observations from the real WM811K analysis and validating the representational quality learned via SSL, Deep TDA, and TL. Minor overlaps observed (e.g., some 'Loc'/'Edge-Loc' appearing in 'None' or 'Center' clusters) reflect the inherent ambiguity and visual similarity between certain synthetically generated patterns, which is expected.

Synergistic integration is key to the framework's success. SSL provides the foundation for learning rich visual representations from unlabeled data. Deep TDA complements this by explicitly encoding multi-scale topological and geometric structures, potentially differentiating patterns with similar textures but different shapes or connectivity, which might be missed by standard CNNs alone. TL provides a mechanism to leverage large, diverse pre-training datasets (incorporating both visual and TDA learning) to build powerful, general-purpose feature extractors, enabling efficient zero-shot analysis and adaptation to new tasks with reduced computational cost and data requirements. The development of a distilled model further enhances practical applicability in resource-constrained environments.

While the results are promising, limitations exist. The synthetic datasets, though designed to mimic real patterns, may not capture the full spectrum of real-world noise, artifacts, or complex defect morphologies. The current defect simulations (strokes, contrast changes, particles) are simplified representations. Future work should involve validation on larger, more diverse proprietary manufacturing datasets, potentially integrating physics-informed constraints or more sophisticated defect generation models. Quantitative comparison with other state-of-the-art unsupervised clustering techniques, particularly those also leveraging SSL or geometric deep learning, would further benchmark performance. Exploring the interpretability of the learned TDA features and their contribution to specific cluster separations remains an important direction for enhancing trust and extracting actionable insights (XAI).

Nevertheless, this study demonstrates that combining Deep TDA with SSL and TL provides a powerful, label-efficient, and adaptable framework for unsupervised analysis of complex semiconductor image data. It addresses key challenges related to data labeling, pattern complexity, and model deployment, offering significant potential for advancing automated process monitoring and quality control in semiconductor manufacturing and other high-volume image analytics domains.

**Conclusion**

This paper addressed the critical challenge of analyzing vast, unlabeled image datasets generated during semiconductor manufacturing for defect identification and process monitoring. We introduced an advanced unsupervised clustering framework that uniquely integrates Deep Topological Data Analysis (Deep TDA), Self-Supervised Learning (SSL), and Transfer Learning (TL). By combining SSL's ability to learn rich visual features without labels, Deep TDA's sensitivity to intrinsic structural and topological patterns, and TL's capacity for efficient knowledge transfer from large pre-trained models, our framework offers a robust and scalable solution.

Validation across diverse datasets, including the real-world WM811K and Mixed WM38 benchmarks, as well as controlled synthetic datasets (SPVD, SWED), demonstrated the framework's effectiveness. It successfully performed unsupervised clustering, grouping images based on underlying defect patterns (single and mixed types) and subtle process variations, even in highly imbalanced scenarios and without relying on labels during training. The integration of TL further enabled efficient zero-shot clustering on new datasets using a pre-trained foundational model incorporating both visual and topological insights.

The primary contribution of this work lies in the synergistic combination of these three powerful techniques, tailored for the complexities of semiconductor image analytics. This integrated approach provides a significant step towards more efficient, insightful, and automated quality control, offering substantial potential for improving yield, reducing costs, and accelerating root cause analysis in semiconductor fabrication and related high-volume manufacturing domains facing similar data challenges.